\documentclass[9pt,twocolumn,twoside]{osajnl}

\journal{ol}
\usepackage{units}
\usepackage{rotating}
\usepackage{booktabs}
\usepackage{times}
\usepackage{multirow}
\usepackage{multicol}
\usepackage{float}
\usepackage{xcolor}
\usepackage{graphicx}
\usepackage{subcaption}
\usepackage{float}
\usepackage{verbatim}
\usepackage{standalone}
\usepackage{tikz}
\usepackage{pgfplots}
\usepackage{csvsimple}

\usetikzlibrary{pgfplots.fillbetween}

\setboolean{shortarticle}{true}
\newcommand{\reffig}[1]{Figure~\ref{fig:#1}}
\newcommand{\reftbl}[1]{Table~\ref{tbl:#1}}

\newcommand{\lblfig}[1]{\label{fig:#1}}

\newcommand{\lbltbl}[1]{\label{tbl:#1}}

\RequirePackage{xcolor}

\definecolor{butter1}{RGB}{252, 233,  79}
\definecolor{butter2}{RGB}{237, 212,   0}
\definecolor{butter3}{RGB}{196, 160,   0}
\colorlet{LightButter}{butter1}
\colorlet{Butter}{butter2}
\colorlet{DarkButter}{butter3}

\definecolor{orange1}{RGB}{252, 175,  62}
\definecolor{orange2}{RGB}{245, 121,   0}
\definecolor{orange3}{RGB}{206,  92,   0}
\colorlet{LightOrange}{orange1}
\colorlet{Orange}{orange2}
\colorlet{DarkOrange}{orange3}

\definecolor{chocolate1}{RGB}{233, 185, 110}
\definecolor{chocolate2}{RGB}{193, 125,  17}
\definecolor{chocolate3}{RGB}{143,  89,   2}
\colorlet{LightChocolate}{chocolate1}
\colorlet{Chocolate}{chocolate2}
\colorlet{DarkChocolate}{chocolate3}

\definecolor{chameleon1}{RGB}{138, 226,  52}
\definecolor{chameleon2}{RGB}{115, 210,  22}
\definecolor{chameleon3}{RGB}{ 78, 154,   6}
\colorlet{LightChameleon}{chameleon1}
\colorlet{Chameleon}{chameleon2}
\colorlet{DarkChameleon}{chameleon3}

\definecolor{skyblue1}{RGB}{114, 159, 207}
\definecolor{skyblue2}{RGB}{ 52, 101, 164}
\definecolor{skyblue3}{RGB}{ 32,  74, 135}
\colorlet{LightSkyBlue}{skyblue1}
\colorlet{SkyBlue}{skyblue2}
\colorlet{DarkSkyBlue}{skyblue3}

\definecolor{plum1}{RGB}{173, 127, 168}
\definecolor{plum2}{RGB}{117,  80, 123}
\definecolor{plum3}{RGB}{ 92,  53, 102}
\colorlet{LightPlum}{plum1}
\colorlet{Plum}{plum2}
\colorlet{DarkPlum}{plum3}

\definecolor{scarletred1}{RGB}{239,  41,  41}
\definecolor{scarletred2}{RGB}{204,   0,   0}
\definecolor{scarletred3}{RGB}{164,   0,   0}
\colorlet{LightScarletRed}{scarletred1}
\colorlet{ScarletRed}{scarletred2}
\colorlet{DarkScarletRed}{scarletred3}

\definecolor{aluminium1}{RGB}{238, 238, 236}
\definecolor{aluminium2}{RGB}{211, 215, 207}
\definecolor{aluminium3}{RGB}{186, 189, 182}
\definecolor{aluminium4}{RGB}{136, 138, 133}
\definecolor{aluminium5}{RGB}{ 85,  87,  83}
\definecolor{aluminium6}{RGB}{ 46,  52,  54}

\definecolor{indigo}{RGB}{114,  33, 188}
\definecolor{maroon}{RGB}{103,   7,  72}
\definecolor{turquoise}{RGB}{ 64, 224, 208}
\definecolor{green4}{RGB}{  0, 139,   0}

\title{\Huge{Does Computer Vision Matter for Action?}}

\author[1*]{Brady Zhou}
\author[1,2]{Philipp Kr\"ahenb\"uhl}
\author[1]{Vladlen Koltun}

\affil[1]{Intel Labs}
\affil[2]{University of Texas at Austin}
\affil[*]{Corresponding author: brady.zhou@utexas.edu}

\begin{abstract}
Controlled experiments indicate that explicit intermediate representations help action.
\end{abstract}

\setboolean{displaycopyright}{false}

\begin{document}

\maketitle

\begin{figure}[b!]
    \centering
    \includegraphics[width=0.4\textwidth]{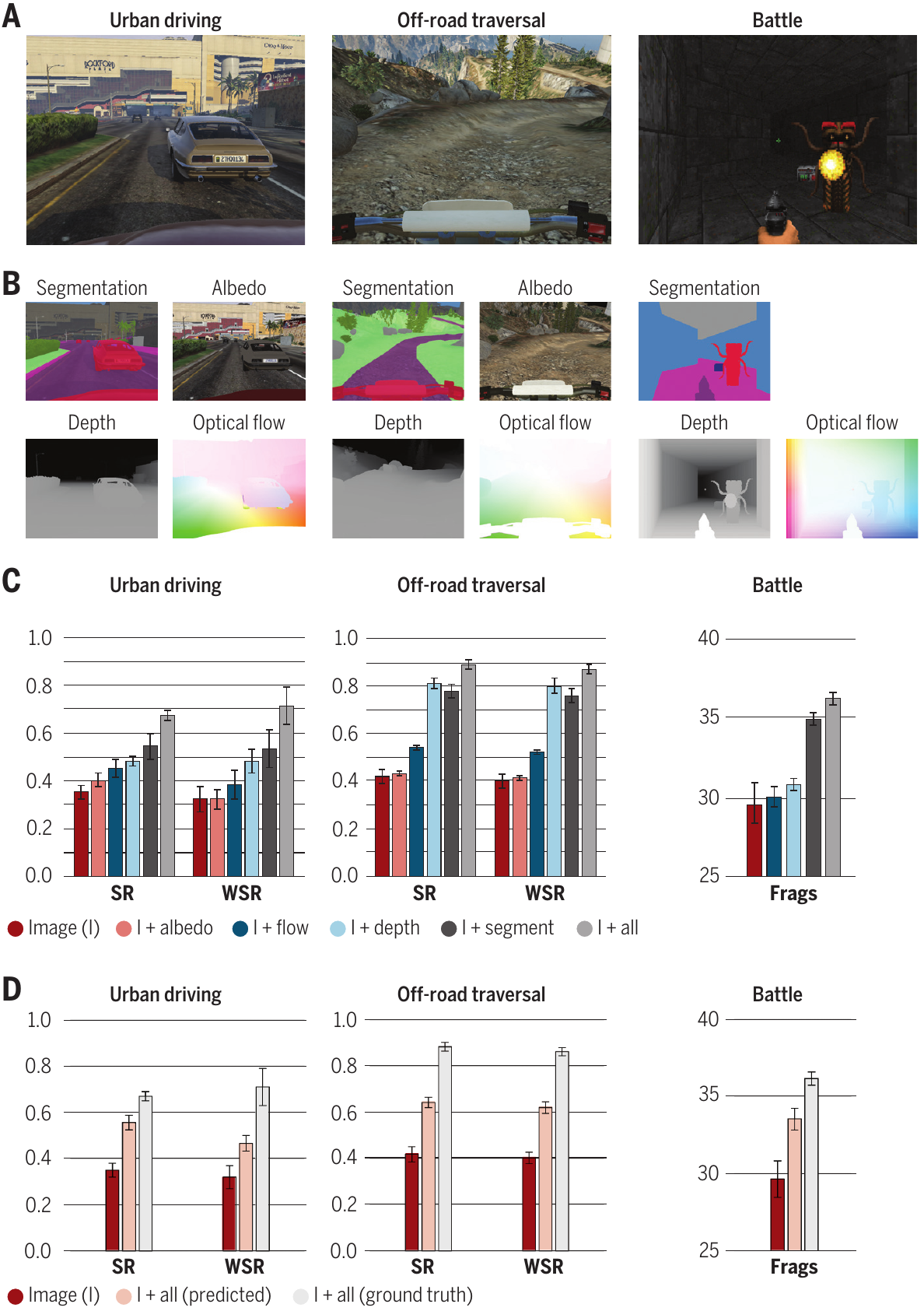}
    \caption{\textbf{Assessing the utility of intermediate representations for sensorimotor control.} (\textbf{A}) Sensorimotor tasks. From left to right: urban driving, off-road trail traversal, and battle. (\textbf{B}) Intermediate representations. Clockwise from top left: semantic segmentation, intrinsic surface color (albedo), optical flow, and depth. (Albedo not used in battle.) (\textbf{C}) Main results. For each task, we compare an image-only agent with an agent that is also provided with ground-truth intermediate representations. The agent observes the intermediate representations during both training and testing. Success rate (`SR') is the fraction of scenarios in which the agent successfully reached the target location; weighted success rate (`WSR') is weighted by track length; `frags' is the number of enemies killed in a battle episode. We show mean and standard deviation in each condition.
(\textbf{D}) Supporting experiments. In the `Image + All (predicted)' condition, the intermediate representations are predicted in situ by a convolutional network; the agent is not given ground-truth representations at test time. The results indicate that even predicted vision modalities confer a significant advantage.
}
    \label{fig:main_fig}
\end{figure}

Biological vision systems evolved to support action in physical environments~\cite{Gibson1979,Churchland1994}. Action is also a driving inspiration for computer vision research. Problems in computer vision are often motivated by their relevance to robotics and their prospective utility for systems that move and act in the physical world. In contrast, a recent stream of research at the intersection of machine learning and robotics demonstrates that models can be trained to map raw visual input directly to action~\cite{Mnih2015,Levine2016,DosovitskiyKoltun2017,Codevilla2018}. These models bypass explicit computer vision entirely. They do not incorporate modules that perform recognition, depth estimation, optical flow, or other explicit vision tasks. The underlying assumption is that perceptual capabilities will arise in the model as needed, as a result of training for specific motor tasks. This is a compelling hypothesis that, if taken at face value, appears to obsolete much computer vision research. If any robotic system can be trained directly for the task at hand, with only raw images as input and no explicit vision modules, what is the utility of further perfecting models for semantic segmentation, depth estimation, optical flow, and other vision tasks?

We report controlled experiments that assess whether specific vision capabilities are useful in mobile sensorimotor systems that act in complex three-dimensional environments. To conduct these experiments, we use realistic three-dimensional simulations derived from immersive computer games. We instrument the game engines to support controlled execution of specific scenarios that simulate tasks such as driving a car, traversing a trail in rough terrain, and battling opponents in a labyrinth. We then train sensorimotor systems equipped with different vision modules and measure their performance on these tasks.

Our baselines are end-to-end pixels-to-actions models that are trained directly for the task at hand. These models do not rely on any explicit computer vision modules and embody the assumption that perceptual capabilities will arise as needed, in the course of learning to perform the requisite sensorimotor task. To these we compare models that receive as additional input the kinds of representations that are studied in computer vision research, such as semantic label maps, depth maps, and optical flow. We can therefore assess whether representations produced in computer vision are useful for sensorimotor challenges. In effect, we ask: What if a given vision task was solved? Would this matter for learning to act in complex three-dimensional environments?

Our first finding is that computer vision does matter. When agents are provided with representations studied in computer vision, they achieve higher performance in sensorimotor tasks. The effect is significant and is consistent across simulation platforms and tasks.

We then examine in finer granularity how useful specific computer vision capabilities are in this context.
Our second finding is that some computer vision capabilities appear to be more impactful for mobile sensorimotor operation than others. Specifically, depth estimation and semantic scene segmentation provide the highest boost in task performance among the individual capabilities we evaluate. Using all capabilities in concert is more effective still.

We also conduct supporting experiments that aim to probe the role of explicit computer vision in detail. We find that explicit computer vision is particularly helpful in generalization, by providing abstraction that helps the trained system sustain its performance in previously unseen environments.
Finally, we show that the findings hold even when agents predict the intermediate representations in situ, with no privileged information.

\section*{Results}

We perform experiments using two simulation environments: the open-world urban and suburban simulation Grand Theft Auto V~\cite{Richter2016,Richter2017,krahenbuhl2018free} and the VizDoom platform for immersive three-dimensional battles~\cite{Kempka2016,DosovitskiyKoltun2017}. In these environments we set up three tasks: urban driving, off-road trail traversal, and battle. The tasks are illustrated in Figure~\ref{fig:main_fig}A.

For each task, we train agents that either act based on the raw visual input alone or are also provided with one or more of the following intermediate representations: semantic and instance segmentation, monocular depth and normals, optical flow, and material properties (albedo). The intermediate representations are illustrated in Figure~\ref{fig:main_fig}B. The environments, tasks, agent architectures, and further details are specified in the supplement.

Figure~\ref{fig:main_fig}C summarizes the main results. Intermediate representations clearly help. The supplement reports additional experiments that examine these findings, possible causes, and alternative hypotheses.

\section*{Analysis}

Our main results indicate that sensorimotor agents can greatly benefit from predicting explicit intermediate representations of scene content, as posited in computer vision research. Across three challenging tasks, an agent that sees not just the image but also the kinds of intermediate representations that are pursued in computer vision research learns significantly better sensorimotor coordination.
Even when the intermediate representations are imperfectly predicted in situ by a small, light-weight deep network, the improvements are significant (Figure~\ref{fig:main_fig}D).

The benefits of explicit vision are particularly salient when it comes to generalization. Equipping a sensorimotor agent with explicit intermediate representations of the scene leads to more general sensorimotor policies. As reported in the supplement, in urban driving, the performance of image-only and image+vision agents is nearly tied on the training set. However, when we test generalization to new areas, the image+vision agent outperforms the image-only agent on the test set even with an order of magnitude less experience with the task during training.

This generalization is exhibited not only by agents equipped with ground-truth representations, but also by agents that predict the intermediate representations in situ, with no privileged information at test time. An agent that explicitly predicts intermediate representations of the scene and uses these explicit representations for control generalizes better to previously unseen test scenarios than an end-to-end pixels-to-actions agent.

Independently of our work, Sax et al.~\cite{sax2018mid} and Mousavian et al.~\cite{mousavian2018visual} studied the role of intermediate representations in visuomotor policies. These works focus on visual navigation in static indoor environments and show that agents equipped with intermediate representations, as studied in computer vision, train faster and generalize better. While the details of environments, tasks, representations, and agents differ, the findings are broadly aligned and appear to support each other.

\section*{Conclusion}

Computer vision produces representations of scene content. Much computer vision research is predicated on the assumption that these intermediate representations are useful for action. Recent work at the intersection of machine learning and robotics calls this assumption into question by training sensorimotor systems directly for the task at hand, from pixels to actions, with no explicit intermediate representations. Thus the central question of our work: Does computer vision matter for action? Our results indicate that it does. Models equipped with explicit intermediate representations train faster, achieve higher task performance, and generalize better to previously unseen environments.

\bibliographystyle{Science}

\section*{Acknowledgments}
{\bf Author contributions.} B.Z., P.K., and V.K. formulated the study, developed the methodology, and designed the experiments. B.Z. and P.K. developed the experimental infrastructure and performed the experiments. B.Z., P.K., and V.K. analyzed data and wrote the paper.
{\bf Competing interests.} The authors declare that they have no competing interests.
{\bf Material availability.} For data not presented in this paper and/or in the supplementary materials, please contact the corresponding author. Code and data for reproducing the results will be released upon publication.

\section*{Supplementary materials}

Supplement S1. Synopsis of findings.

\noindent Supplement S2. Materials and methods.

\noindent Supplement S3. Experiments and analysis.

\newpage

\section*{SUPPLEMENTARY MATERIALS}

\renewcommand{\thesection}{S\arabic{section}}
\renewcommand{\thefigure}{S\arabic{figure}}
\renewcommand{\thetable}{S\arabic{table}}

\pgfplotsset{
	barplot/.style={
		axis x line=bottom,
		axis y line=left,
		ymin=0,
		ymax=1,
		xmin=-0.5,
		xmax=1.5,
		ymajorgrids,
		xtick = {0,1,2,3},
		xticklabels = {SR,WSR,ADT,Rank},
		axis line style={draw=none},
		xtick style={draw=none},
		ytick style={draw=none},
		ybar,
		bar width=5pt,
		legend columns=1,
		legend style={at={(0.5,0.5)},anchor=west},
		width=\textwidth,
		height=6cm,
	}
}

\section{Synopsis of Findings}

We report controlled experiments that assess whether specific vision capabilities, as developed in computer vision research, are useful in mobile sensorimotor systems that act in complex three-dimensional environments. To conduct these experiments, we use realistic three-dimensional simulations derived from immersive computer games. We instrument the game engines to support controlled execution of specific scenarios that simulate tasks such as driving a car, traversing a trail in rough terrain, and battling opponents in a labyrinth. We then train sensorimotor systems equipped with different vision modules and measure their performance on these tasks.

Our baselines are end-to-end pixels-to-actions models that are trained directly for the task at hand. These models do not rely on any explicit computer vision modules and embody the assumption that perceptual capabilities will arise as needed, in the course of learning to perform the requisite sensorimotor task. To these we compare models that receive as additional input the kinds of representations that are produced by modules developed in computer vision research, such as semantic label maps, depth maps, and optical flow. We can therefore assess whether representations produced in computer vision are useful for the sensorimotor challenges we study. In effect, we ask: What if a given vision task was solved? Would this matter for learning to act in complex three-dimensional environments?

Our first finding is that computer vision does matter. When agents are provided with representations studied in computer vision research, such as semantic label maps and depth maps, they achieve higher performance in sensorimotor tasks. The effect is significant and is consistent across simulation platforms and tasks.

We then examine in finer granularity how useful specific computer vision capabilities are in this context. To the extent that the computer vision research community aims to support mobile systems that act in three-dimensional environments, should it invest in improving semantic segmentation accuracy? Optical flow estimation? Intrinsic image decomposition? Do all these tasks matter to the same extent? Our second finding is that some computer vision capabilities appear to be more impactful for mobile sensorimotor operation than others. Specifically, semantic scene segmentation and depth estimation provide the highest boost in task performance among the individual capabilities we evaluate. (Using all capabilities in concert is more effective still.) This can further motivate research on image- and video-based depth estimation, which has gained momentum in recent years, as well as increased investment in solving multiple vision tasks in concert.

We also conduct supporting experiments that aim to probe the role of explicit computer vision in detail. We find that explicit computer vision is particularly helpful in generalization, by providing abstraction that helps the trained system sustain its performance in previously unseen environments. We also find that explicit vision boosts training efficiency, but the performance advantages are maintained even if the baselines are allowed to train ad infinitum. That is, vision does not merely accelerate training, it improves task performance even if training costs are not a factor. We trace this phenomenon to generalization: without the abstraction provided by explicit vision, end-to-end pixels-to-actions models are liable to overfit the training environments. Finally, we show that the findings hold even when agents predict the intermediate representations in situ, with no privileged information.

\section{Materials and Methods}

\subsection*{Environments}
We perform experiments using two simulation environments: The open-world urban and suburban simulation Grand Theft Auto V (GTAV) and the VizDoom platform for immersive three-dimensional battles.

\paragraph{Grand Theft Auto V.}
GTAV is a highly realistic simulation of a functioning city and its surroundings, based on Los Angeles~\cite{Richter2016,Richter2017}.
We instrumented GTAV to extract multiple vision modalities by intercepting and modifying DirectX rendering commands as the game is played~\cite{krahenbuhl2018free}.
The game has an active modding community with tools to access and manipulate game state.
The game's internal scripting interface provides access to control inputs, such as steering, acceleration, and braking of the player's vehicle.
The GTAV environment features a rich set of weather conditions, a full day-night cycle, and a large and diverse game world.
We make use of this during both training and evaluation.
See Figures~\ref{fig:task_driving} and \ref{fig:task_offroad} for examples.

\paragraph{VizDoom.}
The VizDoom environment is a lightweight wrapper around the 1990s video game Doom~\cite{Kempka2016}.
The environment uses a 2.5-dimensional software renderer that stores intermediate representations in memory buffers.
It allows multiple simulations to run in parallel on the same processor, each running at up to $200 \times$ real time.
This makes the VizDoom environment particularly suitable for data-hungry sensorimotor learning algorithms.
VizDoom allows for detailed control over scenarios via a custom scripting interface~\cite{Kempka2016} and level designer.
See \reffig{task_battle} for a visual reference.

\subsection*{Sensorimotor tasks}
We set up three major tasks across the two simulators: urban driving (GTAV), off-road trail traversal (GTAV), and battle (VizDoom).

\paragraph{Urban driving.}
In the urban driving task, an agent controls the continuous acceleration, braking, and steering of a car. The agent needs to navigate a functioning urban road network and avoid obstacles by processing a diverse visual input stream. We use the same car model for all experiments.
An agent starts at one of a set of predefined locations on a street in an urban area.
The goal of the task is to reach a corresponding predefined target location without accidents (e.g., collisions with other vehicles, pedestrians, or the environment).
The start and target locations are directly connected by a major road.
If the route contains intersections, the agent can reach its target by always turning right at intersections.
This avoids higher-level navigation and path finding issues~\cite{Codevilla2018}.
The task ends if the agent comes within 5 meters of the target, or damages the vehicle in any way.
We measure performance on this task via two metrics: success rate (SR) and success rate weighted by track length (WSR).
An agent succeeds in scenario $i$ if it comes within 5 meters of the target; this is indicated by $S_i$.
We report both success rate $\frac{1}{N} \sum_i S_i$ and success rate weighted by track length $\sum_i S_i l_i / \sum_i l_i$, where $l_i$ is the length of track $i$.
Our success rate is analogous to the SPL metric~\cite{Anderson2018}, since we expect the agent to follow a single path from the start location to the target.
Both metrics are normalized, where $0$ means failure on all tracks and $1$ means that the agent reached the target in all tested scenarios.
\reffig{task_driving} shows some input images for the urban driving task.

\paragraph{Off-road trail traversal.}
In this task, the agent operates a dirt bike on a trail in rough terrain. The agent controls acceleration, brake, and steering.
In comparison to urban driving, this task has fewer dynamic obstacles, but is more challenging to navigate. The overall evaluation protocol is analogous to the urban driving task: the agent is tested on a set of scenarios (start/goal locations) and is evaluated in terms of SR and WSR. The goal in each scenario is to reach a goal from a start location.
The agent perceives the world via a head-mounted camera.
\reffig{task_offroad} shows some example input images.

\paragraph{Battle.} This task uses the VizDoom environment.
We follow the task setup of Dosovitskiy and Koltun~\cite{DosovitskiyKoltun2017} (specifically, D3 in their nomenclature).
In each scenario, the goal is to eliminate as many enemies as possible while navigating a labyrinth.
Enemies directly engage and potentially kill the agent.
Secondary goals include the collection of ammunition and health packs.
The agent controls steering and forward/backward/left/right motion, and can fire a weapon.
We measure success by the number of enemies killed (frags) per episode, aggregating statistics over 200,000 time-steps in total.
\reffig{task_battle} shows some example views seen by the agent.

\paragraph{Further details.} For GTAV, we use geographically non-overlapping regions for training, validation, and testing.
An agent never drives or rides through a test area during training.
All parameters are tuned on a separate validation set.
The training set contains $100$ scenarios (start/goal pairs), validation $20$, test $40$.
For VizDoom, we follow the standard experimental setup, using the default parameters of Dosovitskiy and Koltun~\cite{DosovitskiyKoltun2017}, with a train/test split across labyrinth layouts.

\subsection*{Modalities}

We study five input modalities: raw RGB image, monocular depth estimate, semantic and instance segmentation, optical flow, and material estimate (albedo).

\paragraph{RGB image.}
In GTAV, we use the raw RGB image before the head-up display (HUD) is drawn; i.e., without a map, cursor, and other auxiliary information.
We resize the image to $128 \times 128$ pixels and scale the color range to ${[-1,1]}$.
\reffig{modalities_image} shows an example.
In VizDoom, we use the simulator output in grayscale normalized to ${[-1,1]}$, following Dosovitskiy and Koltun~\cite{DosovitskiyKoltun2017}.

\paragraph{Depth.}
We use both the absolute depth estimate obtained directly from the rendering engine and a finite-difference surface normal estimate.
In GTAV, the depth estimate is derived from the depth buffer of the DirectX rendering pipeline.
In VizDoom, the software renderer produces an 8-bit depth estimate.
While the depth image captures the global structure of the scene, the surface normals bring out local detail.
\reffig{modalities_depth} shows an example.

\paragraph{Segmentation.}
We compute both semantic and instance segmentation.
In GTAV, we produce instance-level segmentation for pedestrians, vehicles, and animals, and class-level semantic segmentation for road, sidewalk, terrain, and objects such as signs and trees.
We use the game's scripting engine to track all moving objects, and intercept the rendering commands to segment them~\cite{krahenbuhl2018free}.
For semantic segmentation, we manually labelled all texture IDs in the environment~\cite{Richter2016,Richter2017}.
For VizDoom, we use `item', `enemy', and other object classes, as provided by the simulator.
The label map contains a class-indicator vector for each pixel.
In addition, we use an instance boundary map to represent instance segmentation.
\reffig{modalities_segmentation} shows an example.

\paragraph{Flow.}
We use three different representations of optical flow:
Plain optical flow, as the motion of each pixel between consecutive captured frames;
a static motion estimate, caused by parallax or ego-motion;
a dynamic motion estimate, caused by object motion or deformation.
Static and dynamic motion sum up to optical flow.
Each motion field contains a vertical and horizontal component per pixel.
In GTAV, we track all moving objects using the game's scripting engine to compute the dynamic flow component.
Static flow is derived directly from the 3D location of a pixel and the relative camera motion.
For VizDoom, we track all moving objects as 2D sprites, and compute the static flow component from the depth map and relative camera motion.
We follow the video understanding literature and clip motion estimates into the range ${[-32,32]}$ and divide by $64$ to normalize.
We use backward flow from the current frame to the previous frame, as it is more complete for agents moving forward.
\reffig{modalities_flow} shows an example.

\paragraph{Albedo.}
We compute the intrinsic surface color of each pixel, known as albedo.
The albedo factors out the effect of lighting and represents the intrinsic material properties of objects.
Albedo is represented as an RGB image normalized to ${[-1,1]}$.
This modality is only available in GTAV, where it is explicitly represented in the rendering pipeline.
\reffig{modalities_material} shows an example.

\subsection*{Agents}

We parameterize all agents as reactive deep networks. They observe the current input modalities and predict the next action.

\paragraph{GTAV.}
We train the GTAV agent by imitation learning.
For each task, we construct 100/20/40 scenarios for our train/validation/test sets respectively.
The three sets are located in different geographic regions of the map.
Each scenario ranges from 50 to 400 meters in length, with an average of 250 meters.
To collect training data for imitation learning, we use the game's internal autopilot through the built-in scripting library.
The autopilot has access to the internal game state and completes each scenario perfectly.
The autopilot exposes all the control commands through the scripting interface, but does not allow a player (or agent) to take control once engaged.
We record the autopilot's actions on each training scenario and capture the input modalities, actions (speed, steering angle), and other game state information at 8 frames per second (FPS).
Since the autopilot drives close to perfectly, this basic training scheme never learns to recover from mistakes.
To counteract this, we inject noisy actions following the setup of Codevilla et al.~\cite{Codevilla2018}.
We inject 0.25 seconds of noise for every 5 seconds of driving.
Frames with injected noise are not used for training, the recovery is.

The agent uses a deep fully-convolutional control network $F$ to output the steering angle $\theta$, given the input modalities of the current time step.
We use a network-in-network model~\cite{lin2013network},
with the addition of batch normalization~\cite{ioffe2015batch}
before every major convolution block, and leaky ReLU with slope $0.2$.
These two modifications stabilize and accelerate training.
The architecture is visualized in \reffig{imlearn_arch}.

\begin{figure}[!h]
    \centering
    \includegraphics[width=0.3\textwidth]{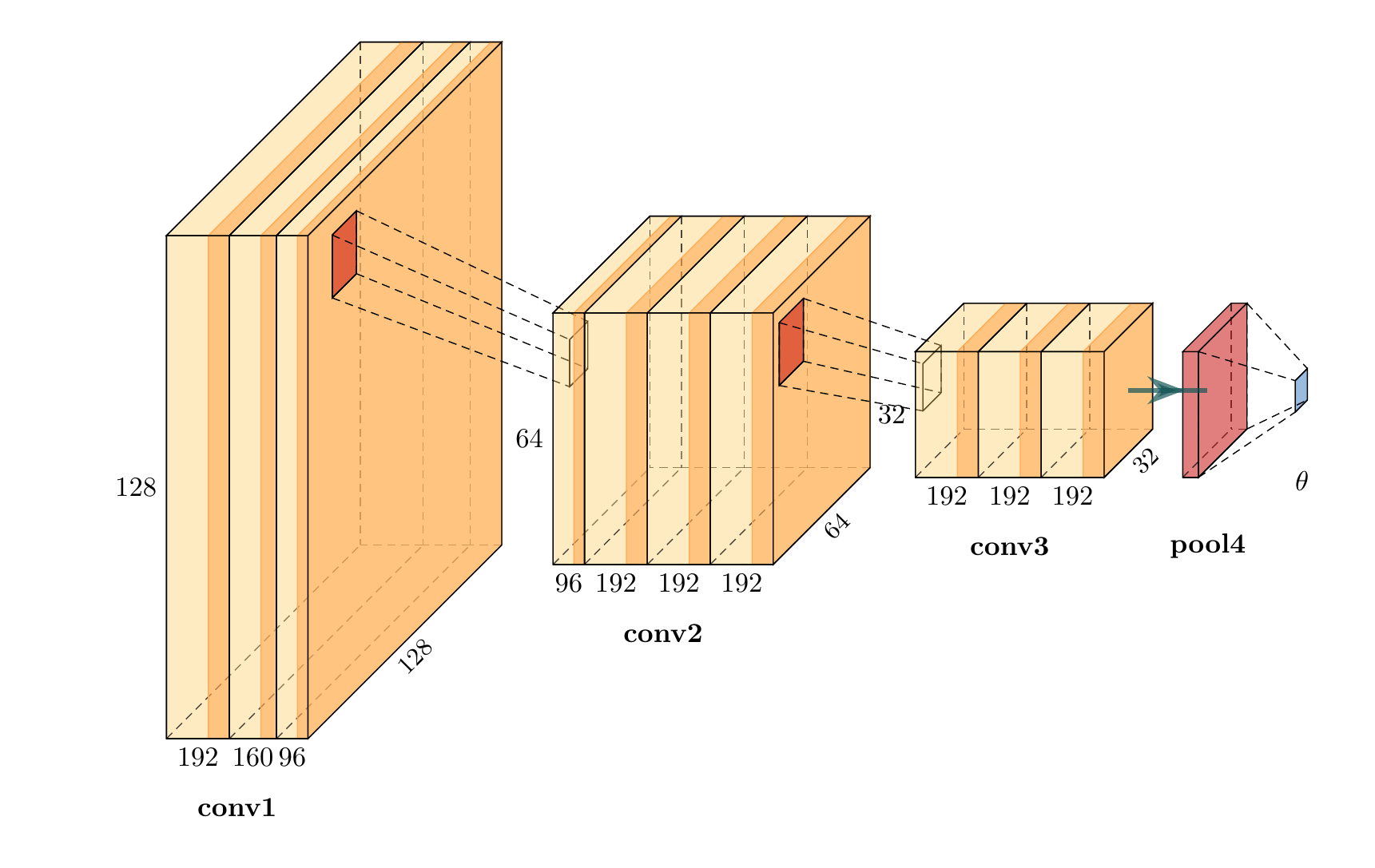}
    \caption{\textbf{Imitation learning agent architecture.}  \reffig{imlearn_arch_detailed} provides more architecture details.}
    \lblfig{imlearn_arch}
\end{figure}
Each agent sees 100K training frames, the equivalent of $3.5$ hours of game time, unless otherwise stated.
Training tracks are visited multiple times under different time-of-day and weather conditions.

The control network is trained using the Adam optimizer~\cite{KingmaBa2015},
with batch size $64$ and learning rate $10^{-4}$, using an $\mathcal{L}_1$ loss between the network output $F(x)$ and the autopilot's output $\theta$.
We experimented with several different loss functions, such as mean squared error or binned angular classification, but the simple $\mathcal{L}_1$ performed best.
We train each network for 100K iterations, decaying the learning rate by $\frac{1}{2}$ every 25k iterations.
Full convergence takes $\sim1$ day.

We use a PID controller for acceleration and braking.
The PID controller targets a constant velocity for each task.
The controller accelerates (positive control) or brakes the vehicle (negative control) to maintain a constant speed of 15 miles per hour for the mountain biking task, and 25 miles per hour for the urban driving task.
The agent updates the control input $8$ times per second.
In between updates the control signal is repeated.

\paragraph{VizDoom.}
The VizDoom agent is trained with Direct Future Prediction (DFP)~\cite{DosovitskiyKoltun2017}, a sensorimotor learning algorithm that uses a stream of measurements as supervisory signal.
A DFP agent takes in three inputs: an image, state measurements (health, ammo, etc.), and a goal vector of the desired state measurements (maximize health, ammo, frags).
A separate neural-network submodule processes each input into an intermediate representation.
A prediction network produces two outputs from this intermediate representation: the expected value of the future measurements over all actions, and the differences in future measurements for all actions.
We use a $6$-layer convolutional network modeled after the Atari architecture of Mnih et al.~\cite{Mnih2015}.
DFP uses an experience buffer starting from a random policy and actions sampled using an $\epsilon$-greedy policy.
$\epsilon$ decays over training epochs.
Given this experience buffer, DFP trains the prediction network by minimizing the $\mathcal{L}_1$ difference between the predictions and the future measurements.
See Dosovitskiy and Koltun~\cite{DosovitskiyKoltun2017} for more details and hyperparameters.

The DFP agent predicts its own future state conditioned on the current action, and chooses the optimal action every four frames.
Each agent sees a total of $50$ million training frames (the equivalent of $2.2$ months of subjective time) unless otherwise stated.
We normalize all input modalities and stack them as channels in the input image.

\subsection*{Vision networks}

To predict the vision modalities, we use a U-net~\cite{ronneberger2015u},
an encoder-decoder architecture with skip connections.
To save computation, we use a single encoder, with a separate decoder for each predicted modality.
The architecture is specified in detail in \reffig{pred_arch}.
The prediction network takes two images as input, representing the current and previous time steps.
(The previous time step is only needed for optical flow.)
We use a separate loss function for each modality.
Depth, normal, factored flow, and albedo use the $\mathcal{L}_1$ distance between the prediction and the ground truth.
We normalize the ground-truth data for training.
Segmentation and boundary prediction are trained using a cross-entropy loss.
All loss functions are summed without weighting them individually.
The network jointly infers all modalities in roughly 100ms.

\begin{figure}[!h]
    \centering
    \includegraphics[width=0.35\textwidth]{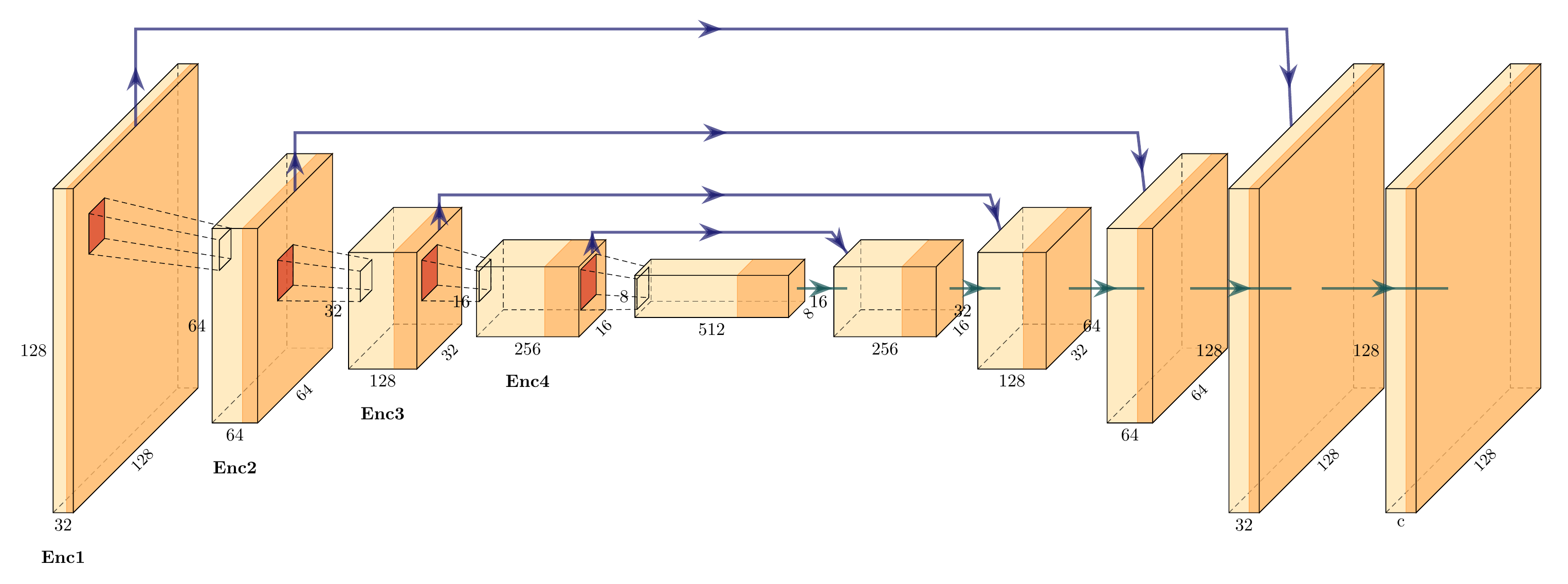}
    \caption{\textbf{Network architecture used to infer the vision modalities.}
    Vision modalities are inferred by a U-net. \reffig{pred_arch_detailed} provides further architecture details.}
    \lblfig{pred_arch}
\end{figure}

\section{Experiments and Analysis}

Figure \ref{fig:main} summarizes the main results on the three tasks.
For each task, we compare an image-only agent with an agent for which various aspects of computer vision are solved.
The agent observes the ground-truth vision modalities during both training and testing.
For GTAV, we report the success rate (SR) and weighed success rate (WSR) across all test scenarios.
We ran the entire evaluation 3 times for different daytime and weather conditions.
We report the mean and standard deviation for each metric.
For VizDoom, we report the average number of enemies killed (frags) over 3 x 50,000 steps of testing.
Here as well we report mean and standard deviation.

The results indicate that ground-truth vision modalities help significantly.
On urban driving and off-road traversal, the success rate roughly doubles when all modalities are used.
In VizDoom, the number of frags increases by more than $20\%$.
All results are statistically significant.

This increased performance could come from various sources. One possibility is that vision-equipped agents generalize better. Another possibility is that the vision modalities provide information that is not otherwise present in the image. We investigate each of these hypotheses independently.

First, we evaluate how much of the vision-equipped agents' performance comes from the ground-truth modalities, and how much can be inferred by a computer vision model that runs in situ with no access to ground-truth data at test time.
This is the `predicted' condition.
\reffig{main_pred} summarizes the results.
Even with a simple U-net model, the inferred vision modalities significantly improve upon the performance of the image-only agent across all tasks.

Next, we investigate if a sufficiently powerful image-only network with the right structure could learn a good generalizable representation given enough data.
We train a sensorimotor agent using the same network structure as the model used in the `predicted' condition, but without ground-truth supervision during training.
This is the `unsupervised' condition.
This agent is trained end-to-end using the same imitation learning objective as the image-only agent, and does not receive any additional supervision.
\reffig{datasize} shows the performance of a ground truth agent, compared to a `predicted' and `unsupervised' agent as the training set size increases on Urban driving and Off-road traversal.
The performance of the unsupervised agent increases with the size of the training set, but requires an order of magnitude more data to reach the performance of the predicted vision agent.

Next, we compare the performance of unsupervised, predicted, and ground-truth agents on the training set. The results are reported in \reftbl{train}. In order to report compatible statistics, we included a subset of the test tracks in the training set for these experiments. \reftbl{train} reports performance on these tracks. All agents attain nearly perfect performance in this condition. This indicates that the differences between these agents in other experiments can be attributed to different generalization abilities.

Finally, \reffig{datasize_battle} shows the performance for the Battle task as the training set grows.
We compare image-only and image+vision agents.
As the training set grows, the image+vision agent quickly begins to outperform the image-only agent, reaching higher performance with less data.

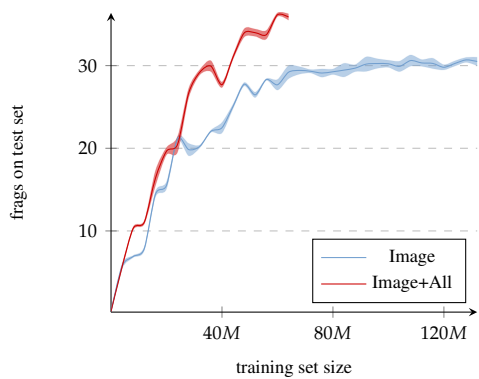
\begin{figure}[h]
    \centering
    \hspace*{-.05\textwidth}
      \begin{tikzpicture}
		\begin{axis}[
			xlabel={\footnotesize training set size},
			ylabel={\footnotesize frags on test set},
			axis x line=bottom,
			axis y line=left,
			xtick={5,10,15},
			xticklabels={$40M$, $80M$, $120M$},
			xticklabel style={font=\footnotesize},
			yticklabel style={font=\footnotesize},
			legend pos=south east,
			ymajorgrids=true,
			grid style=dashed,
			width=0.35\textwidth,
			legend style={font=\footnotesize}
        ]

            \addplot[skyblue1,smooth] coordinates {(0.0, 0.2696) (0.5, 5.7388) (1.0, 6.9068) (1.5, 7.8695) (2.0, 14.4255) (2.5, 15.5161) (3.0, 21.1907) (3.5, 19.8411) (4.0, 20.2267) (4.5, 22.0531) (5.0, 22.4884) (5.5, 25.075) (6.0, 27.7414) (6.5, 26.4677) (7.0, 28.318) (7.5, 27.6828) (8.0, 29.2034) (8.5, 29.4194) (9.0, 29.3113) (9.5, 29.1019) (10.0, 29.258) (10.5, 29.4572) (11.0, 29.6945) (11.5, 30.183) (12.0, 30.2515) (12.5, 30.2199) (13.0, 29.9097) (13.5, 30.6111) (14.0, 30.3264) (14.5, 30.2639) (15.0, 29.7986) (15.5, 30.2657) (16.0, 30.691) (16.5, 30.5)};
 \addlegendentry{Image};
             \addplot[scarletred2,smooth] coordinates {(0.0, 0.2696) (0.5, 5.6823) (1.0, 10.3162) (1.5, 11.0604) (2.0, 16.3654) (2.5, 19.6474) (3.0, 20.4844) (3.5, 26.6846) (4.0, 29.1967) (4.5, 29.9367) (5.0, 27.6936) (5.5, 31.0433) (6.0, 33.95) (6.5, 33.98) (7.0, 33.7449) (7.5, 36.1535) (8.0, 35.8715)};
\addlegendentry{Image+All};
            \addplot[name path=i_top,color=skyblue1,smooth,opacity=0.1] coordinates {(0.0, 0.3601) (0.5, 6.0361) (1.0, 7.0333) (1.5, 8.0911) (2.0, 14.7622) (2.5, 15.946) (3.0, 21.4297) (3.5, 20.5879) (4.0, 20.396) (4.5, 22.1865) (5.0, 23.1238) (5.5, 25.3392) (6.0, 27.9478) (6.5, 26.7935) (7.0, 28.4206) (7.5, 28.3243) (8.0, 29.99) (8.5, 29.8666) (9.0, 29.5213) (9.5, 29.5987) (10.0, 29.5925) (10.5, 30.2796) (11.0, 30.1159) (11.5, 31.0613) (12.0, 30.8296) (12.5, 30.6532) (13.0, 30.5414) (13.5, 31.2886) (14.0, 30.7331) (14.5, 30.8924) (15.0, 30.113) (15.5, 30.5789) (16.0, 30.9906) (16.5, 31.0087)};
            \addplot[name path=i_bot,color=skyblue1,smooth,opacity=0.1] coordinates {(0.0, 0.1791) (0.5, 5.4416) (1.0, 6.7803) (1.5, 7.6479) (2.0, 14.0887) (2.5, 15.0862) (3.0, 20.9518) (3.5, 19.0944) (4.0, 20.0573) (4.5, 21.9197) (5.0, 21.8529) (5.5, 24.8109) (6.0, 27.535) (6.5, 26.1419) (7.0, 28.2153) (7.5, 27.0412) (8.0, 28.4168) (8.5, 28.9721) (9.0, 29.1014) (9.5, 28.6051) (10.0, 28.9236) (10.5, 28.6348) (11.0, 29.273) (11.5, 29.3046) (12.0, 29.6735) (12.5, 29.7866) (13.0, 29.278) (13.5, 29.9336) (14.0, 29.9197) (14.5, 29.6354) (15.0, 29.4842) (15.5, 29.9525) (16.0, 30.3913) (16.5, 29.9913)};
            \addplot[skyblue1,fill opacity=0.5] fill between [of=i_top and i_bot];

            \addplot[name path=a_top,color=scarletred2,smooth,opacity=0.1] coordinates {(0.0, 0.3601) (0.5, 5.9469) (1.0, 10.5002) (1.5, 11.2767) (2.0, 17.2792) (2.5, 20.0862) (3.0, 21.4248) (3.5, 27.2521) (4.0, 29.5119) (4.5, 30.5941) (5.0, 28.0321) (5.5, 31.3181) (6.0, 34.4467) (6.5, 34.4214) (7.0, 34.2627) (7.5, 36.3794) (8.0, 36.2558)};
            \addplot[name path=a_bot,color=scarletred2,smooth,opacity=0.1] coordinates {(0.0, 0.1791) (0.5, 5.4178) (1.0, 10.1323) (1.5, 10.8441) (2.0, 15.4515) (2.5, 19.2086) (3.0, 19.544) (3.5, 26.1171) (4.0, 28.8815) (4.5, 29.2793) (5.0, 27.3551) (5.5, 30.7686) (6.0, 33.4533) (6.5, 33.5386) (7.0, 33.2271) (7.5, 35.9275) (8.0, 35.4873)};
            \addplot[scarletred2,fill opacity=0.5] fill between [of=a_top and a_bot];

		\end{axis}
	  \end{tikzpicture}
    \caption{\textbf{Performance of image-only and vision-equipped agents as a function of training set size on battle.} The shaded area shows the standard deviation of the evaluation metric across different evaluation runs.}
    \lblfig{datasize_battle}
\end{figure}

\begin{table*}[t]
    \centering
    \begin{tabular}{ l c c c c}
    \toprule
     & \multicolumn{2}{c}{Urban driving} & \multicolumn{2}{c}{Off-road traversal} \\
     & SR & WSR & SR & WSR \\
    \midrule
		Unsupervised & 0.97  $\pm$ 0.05 & 0.98  $\pm$ 0.04 & 1.0  $\pm$ 0.0  & 1.0  $\pm$ 0.0      \\
		Predicted    & 1.0   $\pm$ 0.0  & 1.0   $\pm$ 0.0  & 0.94 $\pm$ 0.01  & 0.95 $\pm$ 0.01  \\
		Ground Truth & 0.97  $\pm$ 0.05 & 0.95  $\pm$ 0.08 & 0.98 $\pm$ 0.03 & 0.99 $\pm$ 0.02  \\
    \bottomrule
    \end{tabular}
    \caption{\textbf{Performance of image-only and vision-equipped agents on training tracks. All models fit the training data almost perfectly.}}
    \lbltbl{train}
\end{table*}

\subsection*{Discussion}

Our main results indicate that sensorimotor agents can greatly benefit from predicting explicit intermediate representations of scene content, as posited in computer vision research. Across three challenging tasks, an agent that sees not just the image but also the kinds of intermediate representations that are pursued in computer vision performs significantly better at sensorimotor control.

\paragraph{Which modalities help most?}
Across all tasks, semantic and instance segmentation help significantly.
Semantic segmentation clearly highlights the drivable or walkable area, and obstacles in the agent's path show up in the semantic label map and the instance boundaries.
The drivable area in particular can be challenging to infer from the raw image in an urban environment, while being clearly delineated in the semantic segmentation.
The size and shape of instances in the instance segmentation provides useful cues about the distance and heading of obstacles and other agents.

Depth and normal maps also boost sensorimotor performance.
Nearby obstacles are clearly visible in the depth map and the normal map.
Depth also provides a direct distance estimate to obstacles, allowing the agent to prioritize which scene elements to focus on.

Optical flow and intrinsic material properties appear less helpful.

\paragraph*{Agents with explicit vision generalize better.}
The benefits of explicit vision are particularly salient when it comes to generalization.
Equipping a sensorimotor agent with explicit intermediate representations of the scene leads to more general sensorimotor policies.
In urban driving and off-road traversal, the training set performance of unsupervised, predicted, and ground-truth agents is nearly tied in success rate.
They all fit the training set well, as shown in \reftbl{train}.
However, when we test generalization to new areas, the ground-truth agent outperforms the unsupervised agent on the test set even with an order of magnitude less experience with the task during training (\reffig{datasize}).

\paragraph{Does more data help?}
Could it be that an unsupervised agent simply needs to undergo more training to learn to act well? Perhaps with sufficient training, the performance of the image-only agent will catch up with its vision-equipped counterparts?
Experiments reported in \reffig{datasize} indicate that this is not straightforward.
Even when provided with an order of magnitude more training data, the unsupervised agent does not learn to perform as well as the vision-equipped agent.

\begin{figure*}[h]
    \captionsetup[subfigure]{justification=centering}
    \centering
    \begin{subfigure}{0.35\textwidth}
      \centering
      \begin{tikzpicture}
		\begin{axis}[
			xlabel={\footnotesize training set size},
			ylabel={\footnotesize Success Rate on test tracks},
			axis x line=bottom,
			axis y line=left,
			xtick={0,1,2,3,4},
			xticklabels={$10k$, $30k$, $100k$, $300k$, $1M$},
			xticklabel style={font=\footnotesize},
			xmin=0, xmax=4,
			ymin=0, ymax=0.91,
			ytick={0,0.25,0.5,0.75,1},
			yticklabel style={font=\footnotesize},
			legend pos=south east,
			ymajorgrids=true,
			grid style=dashed,
			width=0.95\textwidth,
			legend style={font=\footnotesize}
		]
		\addplot[chocolate2,smooth] coordinates { (0,0.187)(1,0.248)(2,0.462)(3,0.547)(4,0.573) };
		\addlegendentry{unsupervised};
		\addplot[butter1,smooth] coordinates { (0,0.291)(1,0.385)(2,0.556) };
		\addlegendentry{predicted};
        \addplot[scarletred2,smooth] coordinates { (0, 0.376) (1, 0.53) (2, 0.67) };
		\addlegendentry{ground truth};
		\addplot[name path=u_top,color=chocolate2!20,smooth] coordinates { (0,0.199)(1,0.260)(2,0.483)(3,0.579)(4,0.585) };
		\addplot[name path=u_bot,color=chocolate2!20,smooth] coordinates { (0,0.175)(1,0.236)(2,0.441)(3,0.515)(4,0.561) };
		\addplot[chocolate2,fill opacity=0.5] fill between [of=u_top and u_bot];
		\addplot[name path=n_top,color=butter1!20,smooth] coordinates { (0,0.303)(1,0.406)(2,0.588) };
		\addplot[name path=n_bot,color=butter1!20,smooth] coordinates { (0,0.279)(1,0.364)(2,0.524) };
		\addplot[butter1,fill opacity=0.5] fill between [of=n_top and n_bot];
        \addplot[name path=gtcar_top,color=scarletred2!20,smooth] coordinates { (0, 0.45) (1, 0.554) (2, 0.69) };
        \addplot[name path=gtcar_bot,color=scarletred2!20,smooth] coordinates { (0, 0.302) (1, 0.506) (2, 0.65) };
        \addplot[scarletred2,fill opacity=0.5] fill between [of=gtcar_top and gtcar_bot];
		\end{axis}
	  \end{tikzpicture}
      \caption{Urban driving}
      \lblfig{datasize_urban}
    \end{subfigure}%
    \centering
    \begin{subfigure}{0.35\textwidth}
      \centering
      \begin{tikzpicture}
		\begin{axis}[
			xlabel={\footnotesize training set size},
			ylabel={\footnotesize Success Rate on test tracks},
			axis x line=bottom,
			axis y line=left,
			xtick={0,1,2,3,4},
			xticklabels={$10k$, $30k$, $100k$, $300k$, $1M$},
			xticklabel style={font=\footnotesize},
			xmin=0, xmax=4,
			ymin=0, ymax=0.91,
			ytick={0,0.25,0.5,0.75,1},
			yticklabel style={font=\footnotesize},
			ymajorgrids=true,
			grid style=dashed,
			width=0.95\textwidth,
		]
		\addplot[chocolate2,smooth] coordinates { (0,0.209)(1,0.310)(2,0.519)(3,0.589)(4,0.620) };
		\addplot[butter1,smooth] coordinates { (0,0.295)(1,0.450)(2,0.643) };

		\addplot[name path=u_top,color=chocolate2!20,smooth] coordinates { (0,0.228)(1,0.321)(2,0.559)(3,0.600)(4,0.642) };
		\addplot[name path=u_bot,color=chocolate2!20,smooth] coordinates { (0,0.190)(1,0.299)(2,0.479)(3,0.578)(4,0.598) };
		\addplot[chocolate2,fill opacity=0.5] fill between [of=u_top and u_bot];

		\addplot[name path=n_top,color=butter1!20,smooth] coordinates { (0,0.317)(1,0.461)(2,0.665) };
		\addplot[name path=n_bot,color=butter1!20,smooth] coordinates { (0,0.273)(1,0.439)(2,0.621) };
		\addplot[butter1,fill opacity=0.5] fill between [of=n_top and n_bot];

        \addplot[scarletred2,smooth] coordinates { (0, 0.411) (1, 0.589) (2, 0.884) };
        \addplot[name path=gtbike_top,color=scarletred2!20,smooth] coordinates { (0, 0.451) (1, 0.6) (2, 0.903) };
        \addplot[name path=gtbike_bot,color=scarletred2!20,smooth] coordinates { (0, 0.371) (1, 0.578) (2, 0.865) };
        \addplot[scarletred2,fill opacity=0.5] fill between [of=gtbike_top and gtbike_bot];

		\end{axis}
	  \end{tikzpicture}
      \centering
      \caption{Off-road traversal}
      \lblfig{datasize_urban}
    \end{subfigure}
    \centering
    \caption{\textbf{Performance of unsupervised, predicted, and ground-truth vision agents as a function of training set size.} The shaded area shows the standard deviation of the evaluation metric across different evaluation runs. All models share the same control network architecture. Unsupervised and predicted conditions share the same vision network architecture.}
    \lblfig{datasize}
\end{figure*}
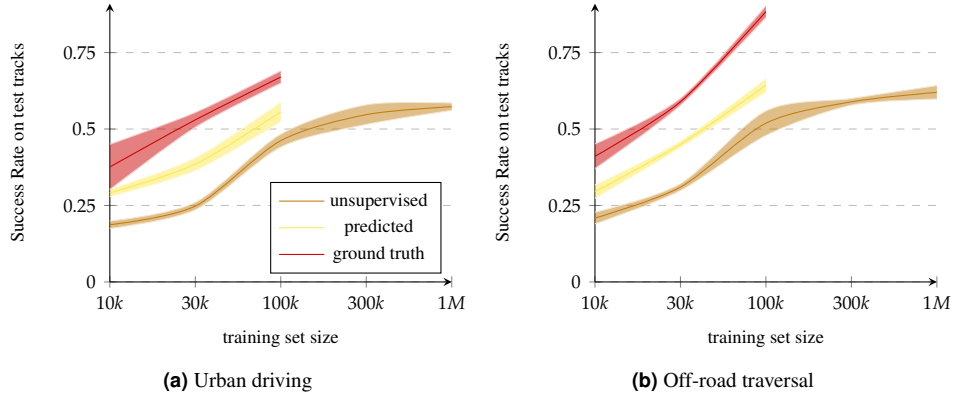

\newpage

\begin{figure*}[!hp]
\captionsetup[subfigure]{justification=centering}
\begin{subfigure}{0.99\textwidth}
\includegraphics[width=0.33\linewidth,page=1]{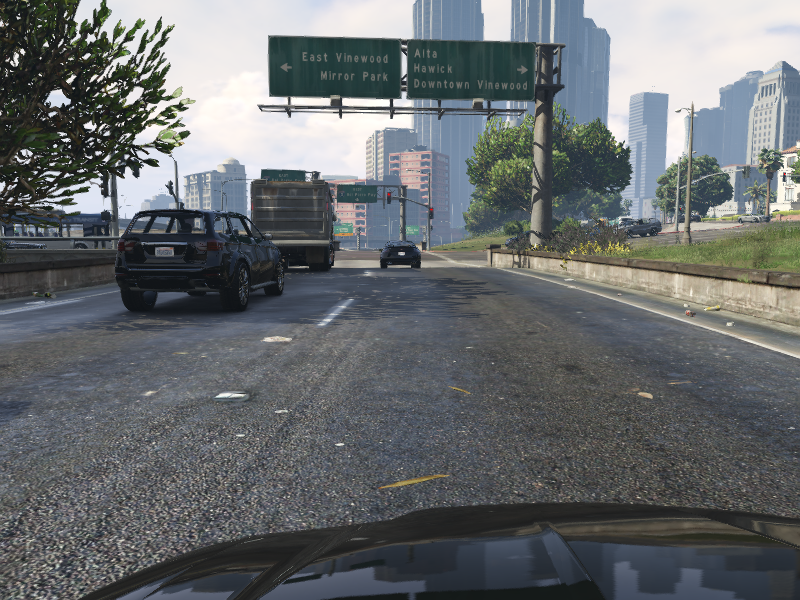}\hspace{0.5mm}%
\includegraphics[width=0.33\linewidth,page=1]{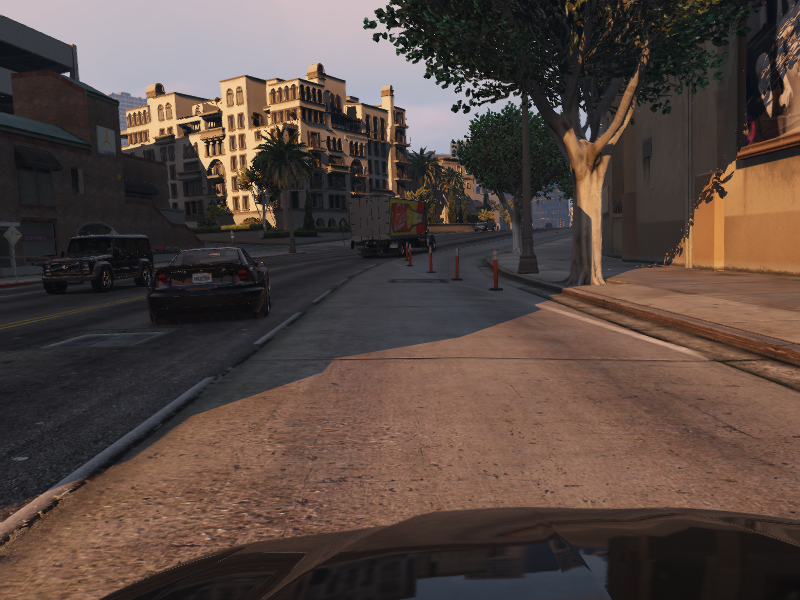}\hspace{0.5mm}%
\includegraphics[width=0.33\linewidth,page=1]{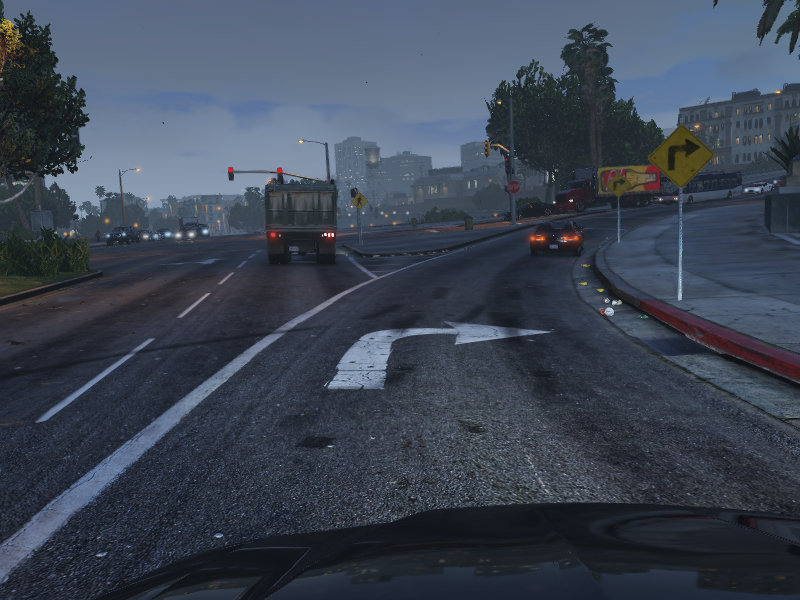}%
\caption{Urban driving}
\lblfig{task_driving}
\end{subfigure}
\begin{subfigure}{0.99\textwidth}
\includegraphics[width=0.33\linewidth,page=2]{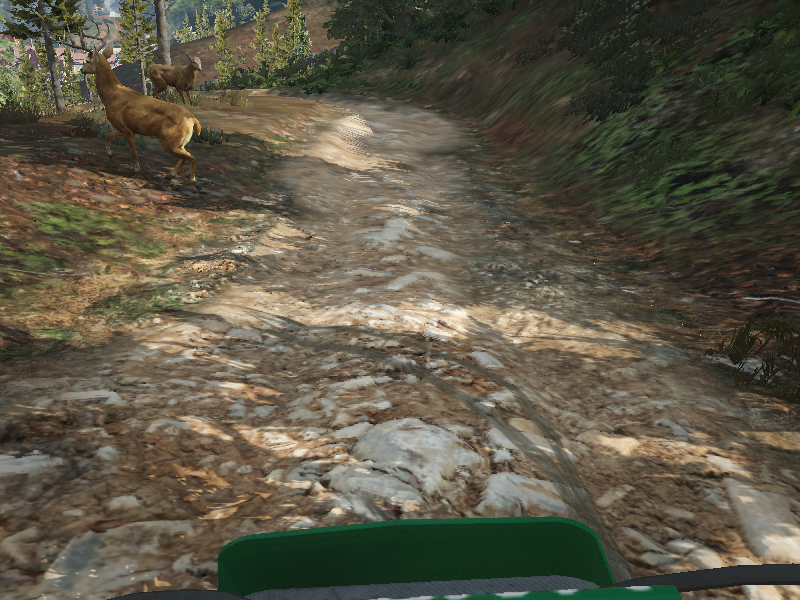}\hspace{0.5mm}%
\includegraphics[width=0.33\linewidth,page=2]{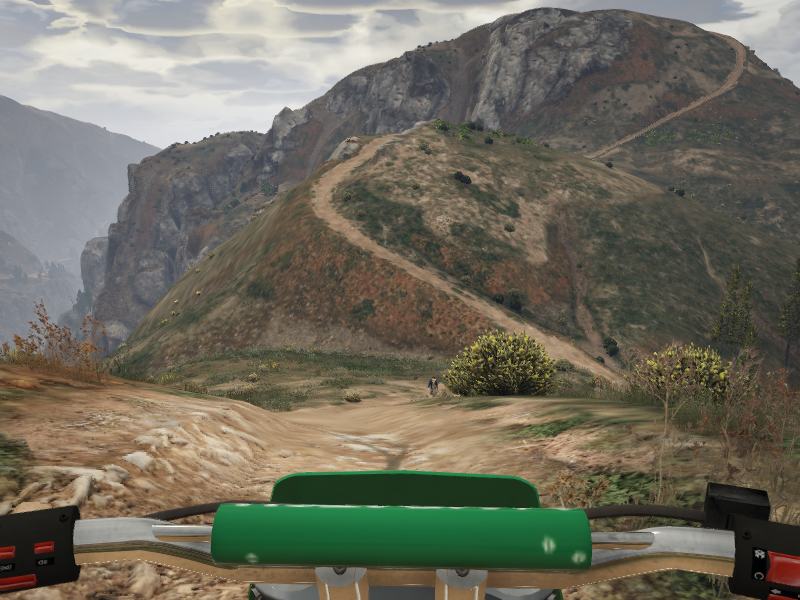}\hspace{0.5mm}%
\includegraphics[width=0.33\linewidth,page=2]{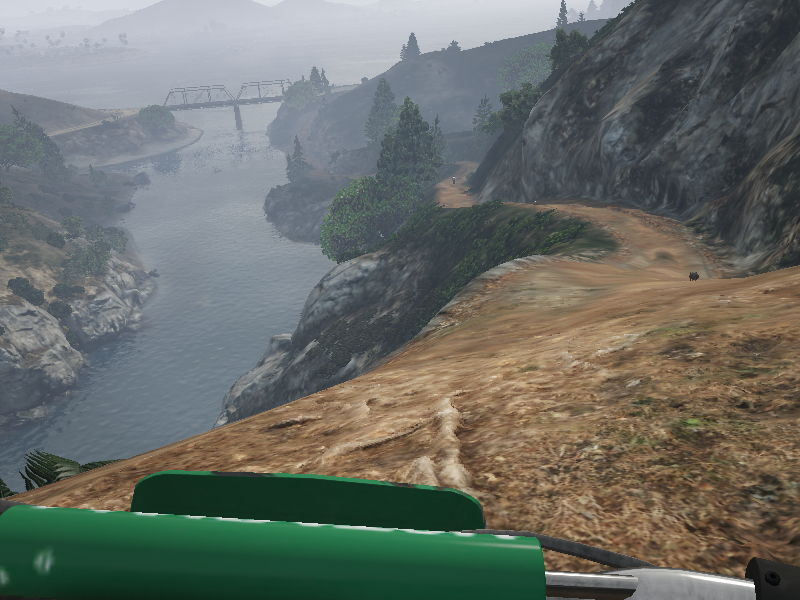}%
\caption{Off-road traversal}
\lblfig{task_offroad}
\end{subfigure}
\begin{subfigure}{0.99\textwidth}
\includegraphics[width=0.33\linewidth,page=3]{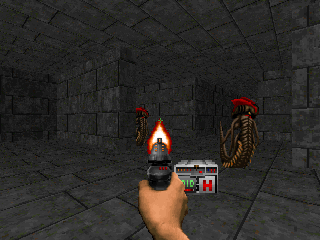}\hspace{0.5mm}%
\includegraphics[width=0.33\linewidth,page=3]{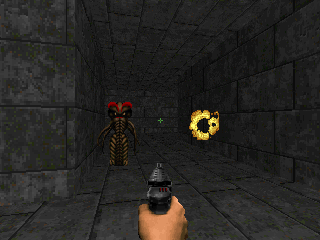}\hspace{0.5mm}%
\includegraphics[width=0.33\linewidth,page=3]{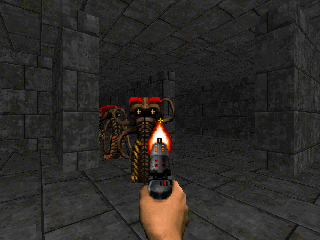}
\caption{Battle}
\lblfig{task_battle}
\end{subfigure}
\caption{\textbf{Three sensorimotor tasks used in our experiments.} For each task, we show three views from the agent's viewpoint.}
\lblfig{task}
\end{figure*}

\newpage
\begin{figure*}[!hp]
\captionsetup[subfigure]{justification=centering}
\begin{subfigure}{0.33\textwidth}
\centering
\includegraphics[width=\linewidth]{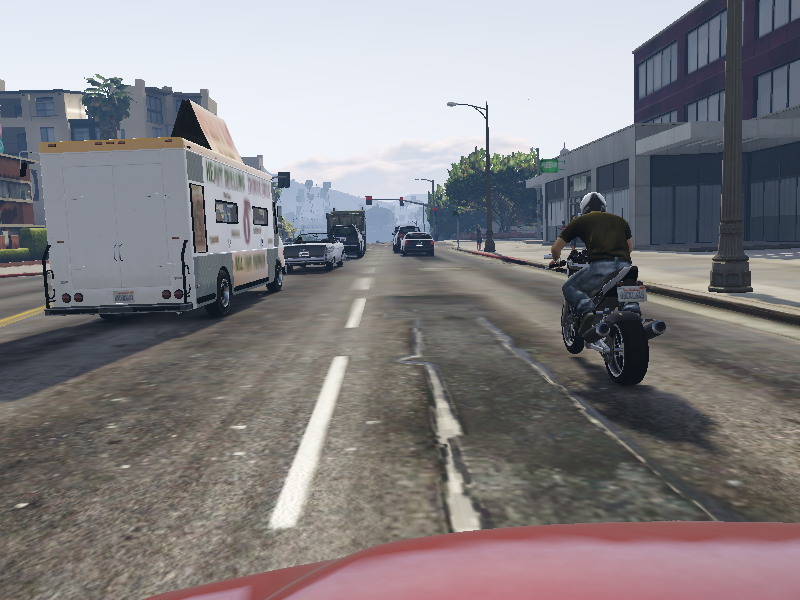}
\caption{RGB Image}
\lblfig{modalities_image}
\end{subfigure}
\begin{subfigure}{0.66\textwidth}
\centering
\includegraphics[width=0.5\linewidth]{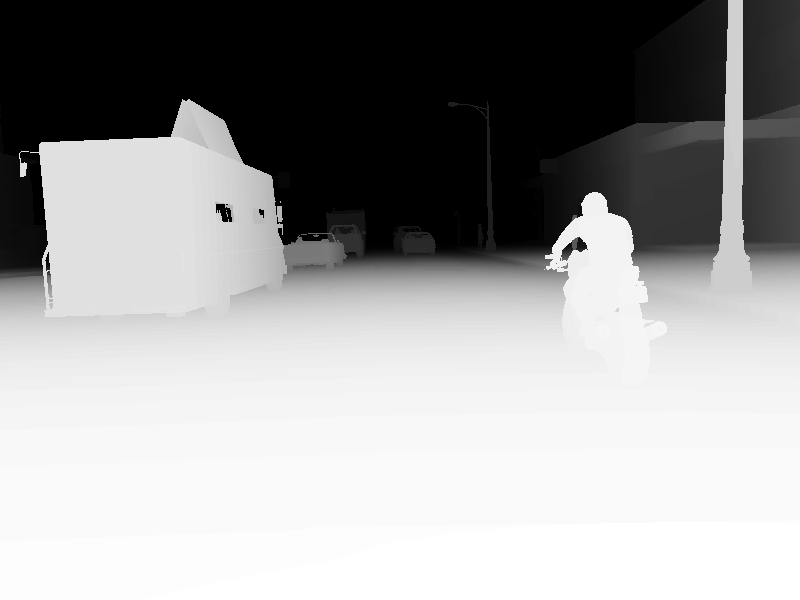}%
\begin{tikzpicture}
\node[inner sep=0  , anchor=north east] () {\includegraphics[width=0.5\linewidth]{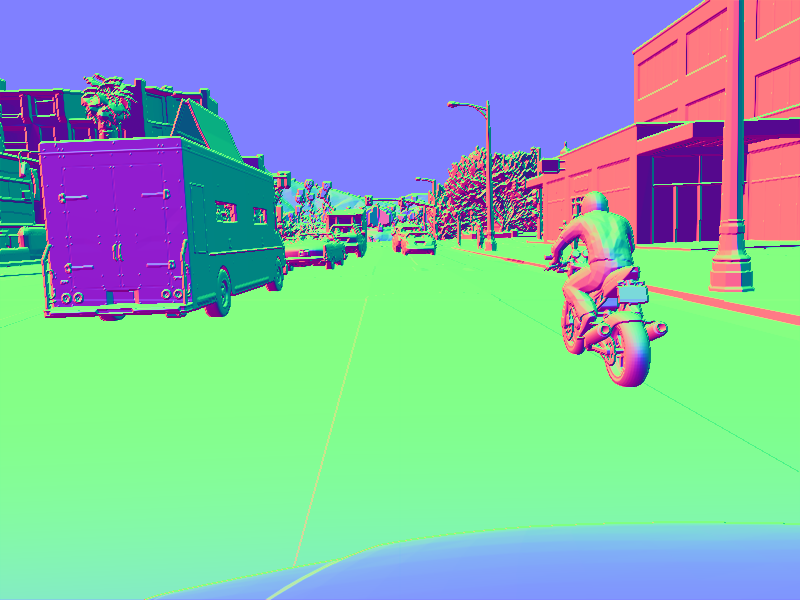}};
\node[inner sep=1pt, anchor=north east, fill=white,draw=black,rounded corners] () {\includegraphics[width=0.1\linewidth]{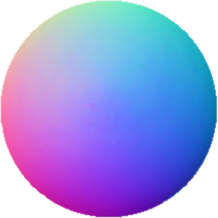}};
\end{tikzpicture}%
\caption{Depth (left) and surface normals (right)}
\lblfig{modalities_depth}
\end{subfigure}
\begin{subfigure}{0.66\textwidth}
\centering
\includegraphics[width=0.5\linewidth,page=4]{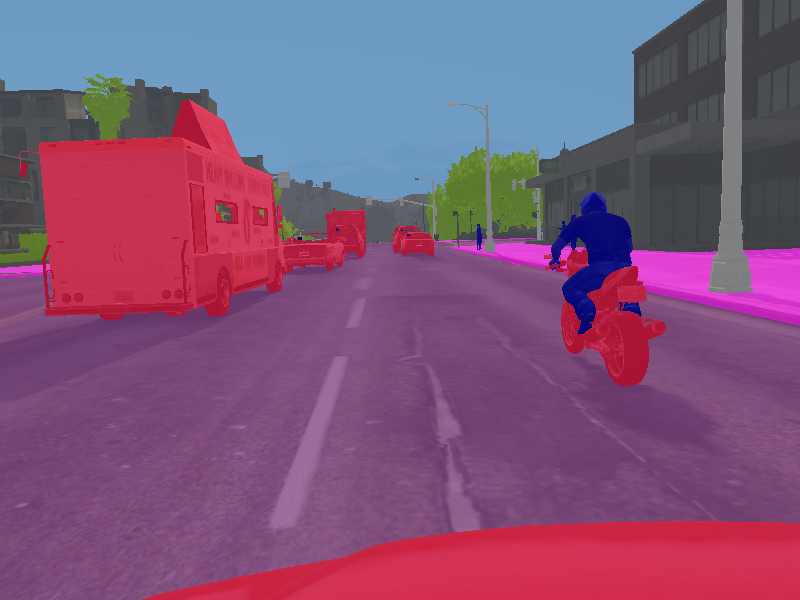}%
\includegraphics[width=0.5\linewidth,page=5]{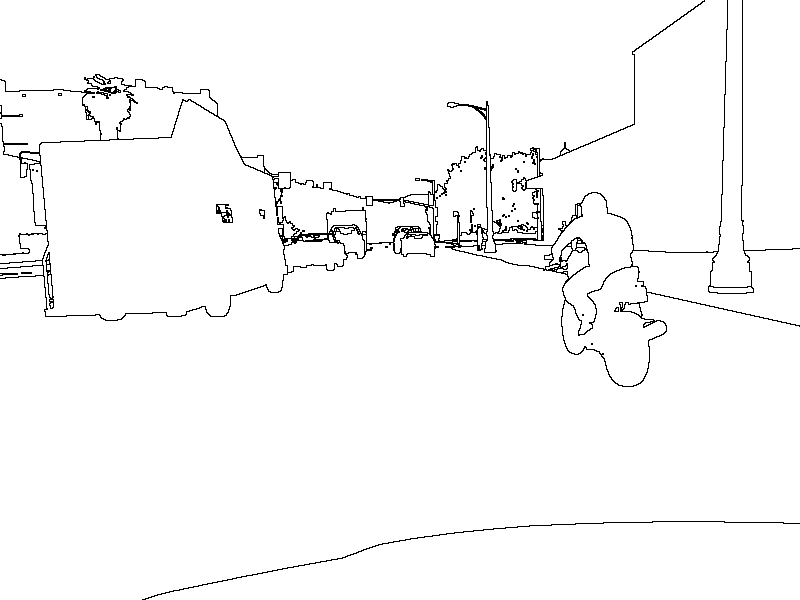}%
\caption{Segmentation: semantic (left) and instance boundaries (right).}
\lblfig{modalities_segmentation}
\end{subfigure}
\begin{subfigure}{0.33\textwidth}
\centering
\includegraphics[width=\linewidth,page=9]{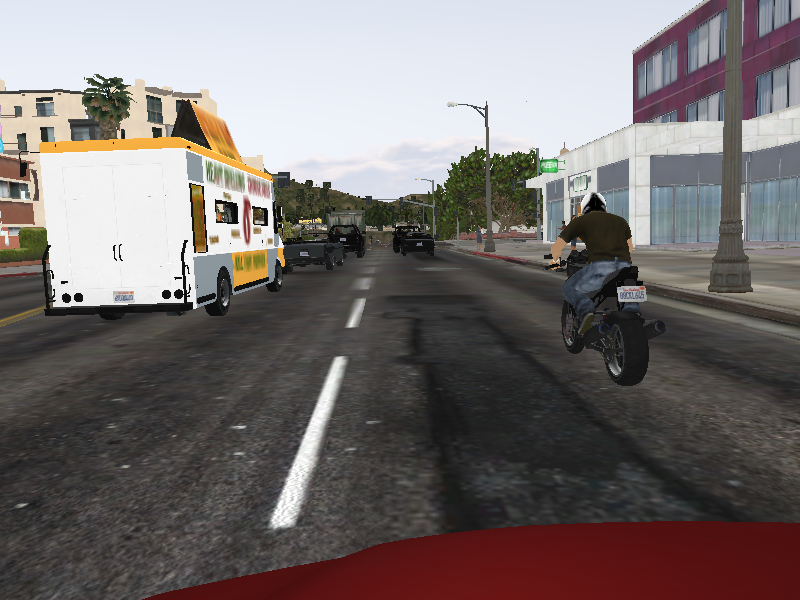}
\caption{Albedo}
\lblfig{modalities_material}
\end{subfigure}
\begin{subfigure}{\textwidth}
\centering
\includegraphics[width=0.33\linewidth,page=6]{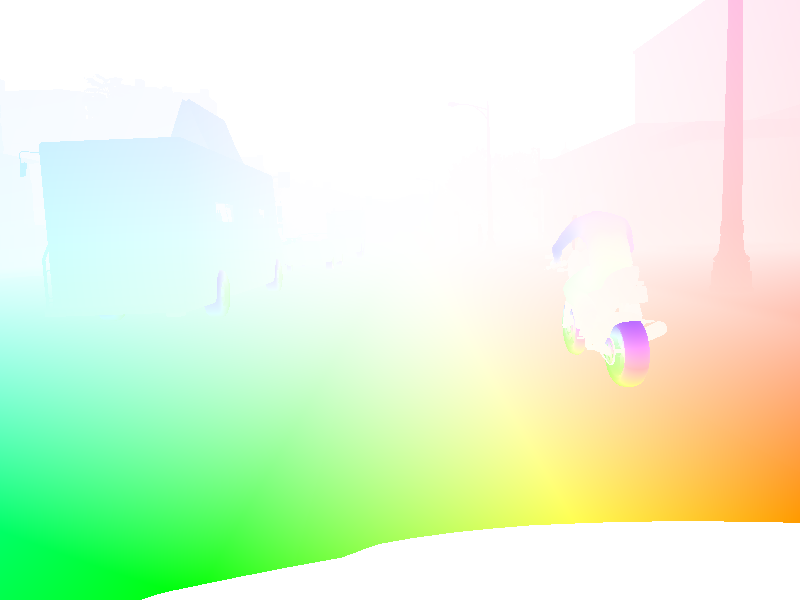}%
\includegraphics[width=0.33\linewidth,page=7]{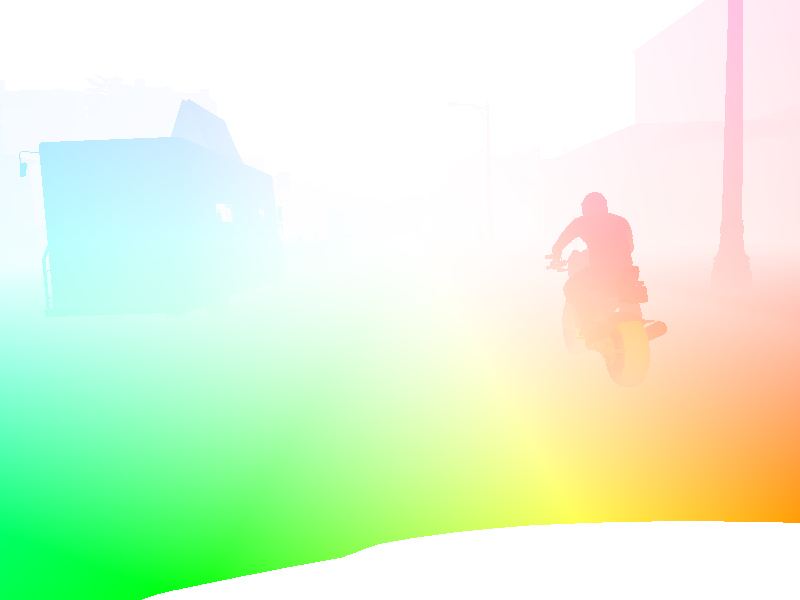}%
\begin{tikzpicture}
\node[inner sep=0  , anchor=north east] () {\includegraphics[width=0.33\linewidth,page=8]{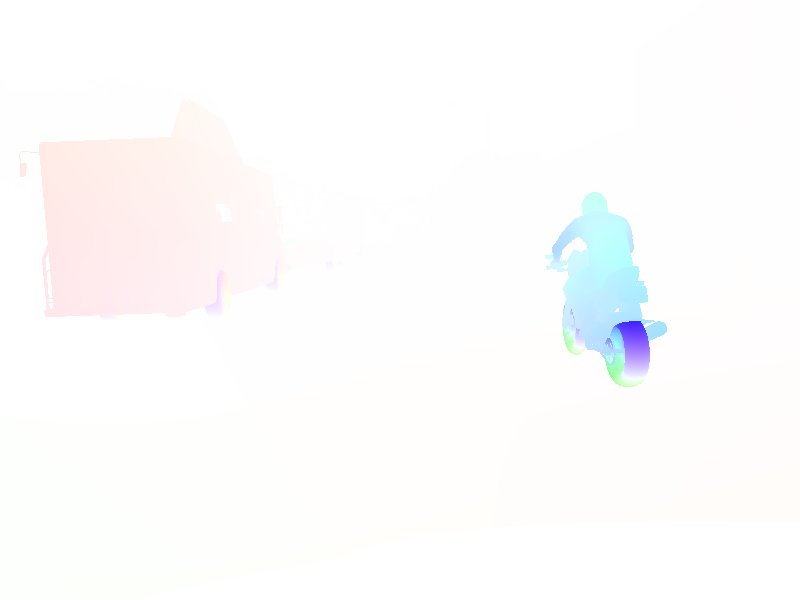}};
\node[inner sep=1pt, anchor=north east, fill=white,draw=black,rounded corners] () {\includegraphics[width=0.066\linewidth]{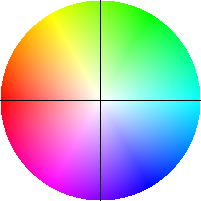}};
\end{tikzpicture}%
\caption{Optical Flow. Full (left) and factored into static (center) and dynamic flow (right).}
\lblfig{modalities_flow}
\end{subfigure}

\caption{\textbf{Different computer vision modalities used in our experiments, illustrated on the urban driving task.} For normal maps, the inset shows the different normal directions projected onto a virtual sphere. For optical flow, the inset shows the flow direction as an offset to the center pixel.}
\lblfig{modalities}
\end{figure*}

\begin{figure*}[!t]
    \captionsetup[subfigure]{justification=centering}
    \begin{subfigure}{0.33\textwidth}
      \centering
		\begin{tikzpicture}
		\begin{axis}[barplot]
		\addplot+[black, fill=skyblue1   ,error bars/.cd, y dir=both, y explicit] coordinates {(0,0.35) +- (0.03,0.03) (1,0.32) +- (0.05,0.05)};
		\addplot+[black, fill=orange1    ,error bars/.cd, y dir=both, y explicit] coordinates {(0,0.40) +- (0.03,0.03) (1,0.32) +- (0.04,0.04)};
		\addplot+[black, fill=plum1      ,error bars/.cd, y dir=both, y explicit] coordinates {(0,0.45) +- (0.04,0.04) (1,0.38) +- (0.06,0.06)};
		\addplot+[black, fill=aluminium3 ,error bars/.cd, y dir=both, y explicit] coordinates {(0,0.48) +- (0.02,0.02) (1,0.48) +- (0.05,0.05)};
		\addplot+[black, fill=chameleon2 ,error bars/.cd, y dir=both, y explicit] coordinates {(0,0.54) +- (0.05,0.05) (1,0.53) +- (0.08,0.08)};
		\addplot+[black, fill=scarletred2,error bars/.cd, y dir=both, y explicit] coordinates {(0,0.67) +- (0.02,0.02) (1,0.71) +- (0.08,0.08)};
		\end{axis}
		\end{tikzpicture}
      \caption{Urban driving}
    \end{subfigure}%
    \captionsetup[subfigure]{justification=centering}
    \begin{subfigure}{0.33\textwidth}
      \centering
		\begin{tikzpicture}
		\begin{axis}[barplot]
		\addplot+[black, fill=skyblue1   ,error bars/.cd, y dir=both, y explicit] coordinates {(0,0.419) +- (0.033,0.033) (1,0.403) +- (0.025,0.025) };
		\addplot+[black, fill=orange1    ,error bars/.cd, y dir=both, y explicit] coordinates {(0,0.434) +- (0.011,0.011) (1,0.409) +- (0.007,0.007) };
		\addplot+[black, fill=plum1      ,error bars/.cd, y dir=both, y explicit] coordinates {(0,0.543) +- (0.011,0.011) (1,0.520) +- (0.009,0.009) };
		\addplot+[black, fill=aluminium3 ,error bars/.cd, y dir=both, y explicit] coordinates {(0,0.806) +- (0.022,0.022) (1,0.797) +- (0.027,0.027) };
		\addplot+[black, fill=chameleon2 ,error bars/.cd, y dir=both, y explicit] coordinates {(0,0.775) +- (0.029,0.029) (1,0.757) +- (0.025,0.025) };
		\addplot+[black, fill=scarletred2,error bars/.cd, y dir=both, y explicit] coordinates {(0,0.884) +- (0.019,0.019) (1,0.863) +- (0.018,0.018) };
		\end{axis}
		\end{tikzpicture}
      \caption{Off-road traversal}
    \end{subfigure}%
    \begin{subfigure}{0.33\textwidth}
      \centering
		\begin{tikzpicture}
		\begin{axis}[barplot,
		  xticklabels = {frags},
		  ymax=40,
		  ymin=25,
		  legend cell align={left},
		]
		\addplot+[black, fill=skyblue1   ,error bars/.cd, y dir=both, y explicit] coordinates {(0,29.64) +- (1.18,1.18)};
		\addlegendimage{black, fill=orange1};
		\addplot+[black, fill=plum1      ,error bars/.cd, y dir=both, y explicit] coordinates {(0,30.01) +- (0.603,0.603)};
		\addplot+[black, fill=aluminium3 ,error bars/.cd, y dir=both, y explicit] coordinates {(0,30.93) +- (0.373,0.373)};
		\addplot+[black, fill=chameleon2 ,error bars/.cd, y dir=both, y explicit] coordinates {(0,34.83) +- (0.417,0.417)};
		\addplot+[black, fill=scarletred2,error bars/.cd, y dir=both, y explicit] coordinates {(0,36.15) +- (0.43,0.43)};
		\legend{Image, Image + Albedo, Image + Flow, Image + Depth, Image + Segm., Image + All}
		\end{axis}
		\end{tikzpicture}
      \caption{Battle}
    \end{subfigure}
		\vspace{3mm}
    \\
    \centering
    \begin{subfigure}{0.75\textwidth}
      {\small
		\begin{tabular}{ cc cccccc }
		\toprule
		& & Image            & I. + Albedo   & I. + Flow     & I. + Depth    & I. + Label    & I. + All     \\
        \midrule
        \multirow{2}{*}{\rotatebox[origin=c]{90}{\parbox[c]{1.5cm}{\centering Urban driving}}}

		& SR   & 0.35 $\pm$ 0.03  & 0.40 $\pm$ 0.03  & 0.45 $\pm$ 0.04  & 0.48 $\pm$ 0.02  & 0.54 $\pm$ 0.05  & 0.67 $\pm$ 0.02 \\[1.25ex]
		& WSR  & 0.32 $\pm$ 0.05  &  0.32 $\pm$ 0.04  &  0.38 $\pm$ 0.06  &  0.48 $\pm$ 0.05  &  0.53 $\pm$ 0.08  &  0.71 $\pm$ 0.08 \\[1.25ex]
		\midrule
		\multirow{2}{*}{\rotatebox[origin=c]{90}{\parbox[c]{1.5cm}{\centering Off-road traversal}}}
		& SR   & 0.42 $\pm$ 0.03  &  0.43 $\pm$ 0.01  &  0.54 $\pm$ 0.01  &  0.81 $\pm$ 0.02  &  0.78 $\pm$ 0.03  &  0.89 $\pm$ 0.02 \\[1.25ex]
		& WSR  & 0.40 $\pm$ 0.03  &  0.41 $\pm$ 0.01  &  0.52 $\pm$ 0.01  &  0.80 $\pm$ 0.03  &  0.76 $\pm$ 0.03  &  0.87 $\pm$ 0.02 \\[1.25ex]
		
		\midrule
		\multirow{3}{*}{\rotatebox[origin=c]{90}{\parbox[c]{1cm}{\centering Battle}}}
		\\
		& Frags& 29.6 $\pm$ 1.2  &    --  & 30.0 $\pm$ 0.6  & 30.9 $\pm$ 0.4  &  34.8 $\pm$ 0.4  &  36.2 $\pm$ 0.4\\
		\\
		\bottomrule
		\end{tabular}}
      \caption{Corresponding numeric results}
    \end{subfigure}

    \caption{\textbf{Performance of agents equipped with different input representations.} The black whiskers show standard deviations across different evaluation runs.}
    \lblfig{main}
\end{figure*}

\begin{figure*}[h!]
    \captionsetup[subfigure]{justification=centering}
    \begin{subfigure}{0.33\textwidth}
      \centering
		\begin{tikzpicture}
		\begin{axis}[barplot]
		\addplot+[black, fill=skyblue1   ,error bars/.cd, y dir=both, y explicit] coordinates {(0,0.35) +- (0.03,0.03) (1,0.32) +- (0.05,0.05)};
		\addplot+[black,fill=orange1,error bars/.cd, y dir=both, y explicit] coordinates {(0,0.453) +- (0.012,0.012) (1,0.316) +- (0.006,0.006)};
		\addplot+[black,fill=plum1,error bars/.cd, y dir=both, y explicit] coordinates {(0,0.436) +- (0.021,0.021) (1,0.334) +- (0.014,0.014)};
		\addplot+[black,fill=aluminium3,error bars/.cd, y dir=both, y explicit] coordinates {(0,0.530) +- (0.012,0.012) (1,0.445) +- (0.022,0.022)};
		\addplot+[black,fill=chameleon2,error bars/.cd, y dir=both, y explicit] coordinates {(0,0.547) +- (0.012,0.012) (1,0.443) +- (0.023,0.023)};
		\addplot+[black,fill=scarletred2,error bars/.cd, y dir=both, y explicit] coordinates {(0,0.556) +- (0.032,0.032) (1,0.466) +- (0.034,0.034)};
		\end{axis}
		\end{tikzpicture}
      \caption{Urban driving}
    \end{subfigure}%
    \begin{subfigure}{0.33\textwidth}
      \centering
		\begin{tikzpicture}
		\begin{axis}[barplot]
		\addplot+[black,fill=skyblue1,error bars/.cd, y dir=both, y explicit] coordinates {(0,0.419) +- (0.033,0.033) (1,0.403) +- (0.025,0.025)};
		\addplot+[black,fill=orange1,error bars/.cd, y dir=both, y explicit] coordinates {(0,0.512) +- (0.033,0.033) (1,0.481) +- (0.035,0.035)};
		\addplot+[black,fill=plum1,error bars/.cd, y dir=both, y explicit] coordinates {(0,0.465) +- (0.019,0.019) (1,0.444) +- (0.017,0.017)};
		\addplot+[black,fill=aluminium3,error bars/.cd, y dir=both, y explicit] coordinates {(0,0.651) +- (0.000,0.000) (1,0.649) +- (0.007,0.007)};
		\addplot+[black,fill=chameleon2,error bars/.cd, y dir=both, y explicit] coordinates {(0,0.643) +- (0.029,0.029) (1,0.630) +- (0.035,0.035)};
		\addplot+[black,fill=scarletred2,error bars/.cd, y dir=both, y explicit] coordinates {(0,0.643) +- (0.022,0.022) (1,0.621) +- (0.025,0.025)};
		\end{axis}
		\end{tikzpicture}
      \caption{Off-road traversal}
    \end{subfigure}%
    \begin{subfigure}{0.33\textwidth}
      \centering
		\begin{tikzpicture}
		\begin{axis}[barplot,
		  xticklabels = {frags},
		  ymax=40,
		  ymin=25,
		  legend cell align={left},
		]
		\addplot+[black, fill=skyblue1   ,error bars/.cd, y dir=both, y explicit] coordinates {(0,29.64) +- (1.18,1.18)};
		\addlegendimage{black, fill=orange1};
		\addplot+[black,fill=plum1,error bars/.cd, y dir=both, y explicit] coordinates {(0,30.35) +- (0.19,0.19)};
		\addplot+[black,fill=aluminium3,error bars/.cd, y dir=both, y explicit] coordinates {(0,30.36) +- (0.18,0.18)};
		\addplot+[black,fill=chameleon2,error bars/.cd, y dir=both, y explicit] coordinates {(0,34.28) +- (0.80,0.80)};
		\addplot+[black,fill=scarletred2,error bars/.cd, y dir=both, y explicit] coordinates {(0,34.00) +- (0.42,0.42)};
		\legend{Image, Image + Albedo, Image + Flow, Image + Depth, Image + Segm., Image + All}
		\end{axis}
		\end{tikzpicture}
      \caption{Battle}
    \end{subfigure}
	\vspace{3mm} \\
	
    \centering
    \begin{subfigure}{0.75\textwidth}
    {\small
    \begin{tabular}{ cc cccccc }
    \toprule
    & & Image & I. + Albedo & I. + Flow & I. + Depth & I. + Label & I. + All \\
    \midrule
    \multirow{2}{*}{\rotatebox[origin=c]{90}{\parbox[c]{1.5cm}{\centering Urban driving}}}
    & SR   &  0.35 $\pm$ 0.03 & 0.45 $\pm$ 0.01 & 0.44 $\pm$ 0.02 & 0.53 $\pm$ 0.01 & 0.55 $\pm$ 0.01 & 0.56 $\pm$ 0.03 \\[1.25ex]
    & WSR  &  0.32 $\pm$ 0.05 & 0.32 $\pm$ 0.01 & 0.33 $\pm$ 0.01 & 0.45 $\pm$ 0.02 & 0.44 $\pm$ 0.02 & 0.47 $\pm$ 0.03 \\[1.25ex]
    \midrule
    \multirow{2}{*}{\rotatebox[origin=c]{90}{\parbox[c]{1.5cm}{\centering Off-road traversal}}}
    & SR   &  0.42 $\pm$ 0.03 & 0.51 $\pm$ 0.03 & 0.47 $\pm$ 0.02 & 0.65 $\pm$ 0.00 & 0.64 $\pm$ 0.03 & 0.64 $\pm$ 0.02 \\[1.25ex]
    & WSR  &  0.40 $\pm$ 0.03 & 0.48 $\pm$ 0.04 & 0.44 $\pm$ 0.02 & 0.65 $\pm$ 0.01 & 0.63 $\pm$ 0.04 & 0.62 $\pm$ 0.03 \\[1.25ex]
    \midrule
    \multirow{3}{*}{\rotatebox[origin=c]{90}{\parbox[c]{1cm}{\centering Battle}}}
    \\
    & Frags& 29.6 $\pm$ 1.2  &  -- & 30.35 $\pm$ 0.19 & 30.36 $\pm$ 0.18 & 34.28 $\pm$ 0.80 & 34.00 $\pm$ 0.42 \\
    \\
    \bottomrule
    \end{tabular}}
    \caption{Corresponding numeric results}
    \end{subfigure}

    \caption{\textbf{Performance of agents equipped with different predicted input representations.} The black whiskers show standard deviations across different evaluation runs.}
    \lblfig{main_pred}
\end{figure*}

\begin{figure*}[h]
    \setlength\extrarowheight{-3pt}
    \centering
    \begin{tabular}{ l c }
        \toprule
        Layer & Output shape \\
        \midrule
        Image & $128 \times 128 \times c$ \\
        \midrule
        Conv 5x5 + LReLU & $128 \times 128 \times 192$ \\
        Conv 1x1 + LReLU & $128 \times 128 \times 160$ \\
        Conv 1x1 + LReLU & $128 \times 128 \times 96$ \\
        Max-Pool 3x3 & $64 \times 64 \times 96$ \\
        Dropout & $64 \times 64 \times 96$ \\
        \midrule
        Batch Norm & $64 \times 64 \times 96$ \\
        Conv 5x5 + LReLU & $64 \times 64 \times 192$ \\
        Conv 1x1 + LReLU & $64 \times 64 \times 192$ \\
        Conv 1x1 + LReLU & $64 \times 64 \times 192$ \\
        Avg-Pool 3x3 & $32 \times 32 \times 192$ \\
        Dropout & $32 \times 32 \times 192$ \\
        \midrule
        Batch Norm & $32 \times 32 \times 192$ \\
        Conv 3x3 + LReLU & $32 \times 32 \times 192$ \\
        Conv 1x1 + LReLU & $32 \times 32 \times 192$ \\
        Conv 1x1 & $32 \times 32 \times 1$ \\
        \midrule
        Global Average Pooling & 1 \\
        \bottomrule
    \end{tabular}
    \caption{\textbf{Imitation learning agent architecture.} The number of input channels is dependent on which vision modalities the agent has access to. The last layer of the network averages the values over all pixel locations to produce the final action.}
    \lblfig{imlearn_arch_detailed}
\end{figure*}

\begin{figure*}[b]
    \setlength\extrarowheight{-3pt}
    \centering
    \captionsetup[subfigure]{justification=centering}
    \begin{subfigure} {0.4\textwidth}
        \centering
        \begin{tabular}{ l c }
            \toprule
            Layer & Output shape \\
            \midrule
            Current + Prev RGB & $128 \times 128 \times 6$ \\
            \midrule
            Conv 3x3 (Encoder 1) & $128 \times 128 \times 32$ \\
            Max-Pool 3x3 & $64 \times 64 \times 32$ \\
            ConvBlock (Encoder 2) & $64 \times 64 \times 64$ \\
            Max-Pool 3x3 & $32 \times 32 \times 64$ \\
            ConvBlock (Encoder 3) & $32 \times 32 \times 128$ \\
            Max-Pool 3x3 & $16 \times 16 \times 128$ \\
            ConvBlock (Encoder 4) & $16 \times 16 \times 256$ \\
            Max-Pool 3x3 & $8 \times 8 \times 256$ \\
            ConvBlock & $8 \times 8 \times 512$ \\
            \bottomrule
        \end{tabular}
        \caption{Encoder architecture.}
    \end{subfigure}
    \captionsetup[subfigure]{justification=centering}
    \begin{subfigure} {0.4\textwidth}
        \centering
        \begin{tabular}{ l c }
            \toprule
            Layer & Output shape \\
            \midrule
            Encoder Features & $8 \times 8 \times 512$ \\
            \midrule
            Conv 3x3 & $8 \times 8 \times 256$ \\
            Upsample & $16 \times 16 \times 256$ \\
            Concat Encoder 4 & $16 \times 16 \times 512$ \\
            ConvBlock & $16 \times 16 \times 256$ \\
            \midrule
            Conv 3x3 & $16 \times 16 \times 128$ \\
            Upsample & $32 \times 32 \times 128$ \\
            Concat Encoder 3 & $32 \times 32 \times 256$ \\
            ConvBlock & $32 \times 32 \times 128$ \\
            \midrule
            Conv 3x3 & $32 \times 32 \times 64$ \\
            Upsample & $64 \times 64 \times 64$ \\
            Concat Encoder 2 & $64 \times 64 \times 128$ \\
            ConvBlock & $64 \times 64 \times 64$ \\
            \midrule
            Conv 3x3 & $64 \times 64 \times 32$ \\
            Upsample & $128 \times 128 \times 32$ \\
            Concat Encoder 1 & $128 \times 128 \times 64$ \\
            ConvBlock & $128 \times 128 \times 32$ \\
            \midrule
            Conv 3x3 & $128 \times 128 \times c$ \\
            \bottomrule
        \end{tabular}
        \caption{Decoder architecture.}
    \end{subfigure}
    \caption{\textbf{Network architecture used to infer the vision modalities.}
    A ConvBlock consists of BN-Conv3x3-LReLU-Conv3x3-LRelu-Conv1x1.
    In the encoder, the ConvBlock output has twice as many channels as the input.
    In the decoder, the ConvBlock outputs have half as many channels as the input.
    One decoder network is trained for each predicted modality. All modalities share the encoder. The depicted structure is used in urban driving and off-road traversal. Battle agents use an analogous structure with one less layer in the encoder and the decoder.}
    \lblfig{pred_arch_detailed}
\end{figure*}

\bibliographystyle{Science}

\end{document}